%% file: Samuel_Dodge_-_ISACS_2013.tex
\documentclass[runningheads,a4paper]{llncs}

\usepackage{amssymb,amsmath}
\usepackage{graphicx,subfig}

\usepackage{pgfplots}
\usetikzlibrary{patterns}
\usepackage{array}

\graphicspath{{./Figures/}}

\usepackage{url}

\newlength{\sw}
\begin{document}

\mainmatter  

\title{Is Bottom-Up Attention Useful for Scene Recognition?}

\titlerunning{Is Bottom-Up Attention Useful for Scene Recognition?}

\author{Samuel F. Dodge \and Lina J. Karam}
\authorrunning{Samuel F. Dodge \and Lina J. Karam}

\institute{Arizona State University\\School of Electrical, Computer and Energy Engineering\\
Tempe, Arizona, USA\\
\path|{sfdodge, karam}@asu.edu|\\
}

\maketitle


\begin{abstract}
The human visual system employs a selective attention mechanism to understand the visual world in an efficient manner. In this paper, we show how computational models of this mechanism can be exploited for the computer vision application of scene recognition. First, we consider saliency weighting and saliency pruning, and provide a comparison of the performance of different attention models in these approaches in terms of classification accuracy. Pruning can achieve a high degree of computational savings without significantly sacrificing classification accuracy. In saliency weighting, however, we found that classification performance does not improve. In addition, we present a new method to incorporate salient and non-salient regions for improved classification accuracy. We treat the salient and non-salient regions separately and combine them using Multiple Kernel Learning. We evaluate our approach using the UIUC sports dataset and find that with a small training size, our method improves upon the classification accuracy of the baseline bag of features approach.
\end{abstract}

\section{Introduction}
When presented with the visual world, the human visual system (HVS) is bombarded with more information than can be processed with its limited resources, and perhaps more information than it actually needs. It is impossible to process all of this information simultaneously. Thus the visual system must selectively choose regions of the scene to focus its attention on. The remaining information is still processed, but at a lower acuity than the attended regions \cite{perif}. This mechanism enables fast scene understanding; the HVS can quickly locate the regions of interest and then parse the scene.

Much work has been done to attempt to model the bottom-up portion of the selectivity mechanism \cite{borji-tip,frintrop-survey}. In these models, low-level image features are pooled in a way that computes saliency by modeling the center surround \cite{itti98,gbvs} or similar mechanism \cite{rare} of the HVS. Recently, these models have begun to see use in computer vision applications \cite{ruti,vigECCV}. Computer vision systems are very much under similar constraints as the HVS, and thus the same kind of selectivity mechanism can prove useful. Most often, saliency is used as a sort of preprocessing stage to reduce the amount of irrelevant information to be processed.

This paper consists of two main parts: an evaluation of models for saliency-based feature pruning and weighting, and a description of a new technique for improving classification performance using saliency. We first evaluate saliency pruning and saliency weighting using five saliency models. Although there are many evaluations of saliency models on predicting eye tracking \cite{borji-tip}, little effort has been made to evaluate different saliency models for a computer vision \emph{recognition} task. However, from this evaluation we conclude that both saliency pruning and weighting do not yield higher classification accuracy because they discard potentially useful information. Given this, we then propose a new method to improve recognition performance. Our approach incorporates both the salient regions and the non-salient regions for classification. We show that this approach yields improved classification accuracy over a baseline bag of words approach.

\section{Bottom-Up Saliency for Descriptor Pruning}


\subsection{Previous Work}
In previous work, saliency has been used as prior information to prune the object search space \cite{ruti}, or prune local features \cite{borji-scene,khan,vigECCV}. The idea here is that what is salient is the most important for recognition, while the non-salient regions consist of distracting information that lowers recognition performance.

In one of the first approaches that incorporates saliency in computer vision, Rutishauser et al. \cite{ruti} use bottom-up saliency to generate a region of interest for use in object recognition. In this region a traditional SIFT-based algorithm is used to recognize the object. The authors show that using saliency gives a higher ROC score, however it is not difficult to imagine situations where the desired object does not fall into the salient area of interest and this method would fail. 

Similarly, Borji and Itti \cite{borji-scene} use saliency to prune dense SIFT-based and C2 features for the application of scene recognition. Although the saliency pruning significantly decreases the number of local features that must be processed, the overall classification accuracy decreases as well. Moreover, the authors use only the Itti model and do not evaluate the performance of other models.

More recently, Khan et al. \cite{khan} use category-specific color saliency to weight, rather than prune, shape-based features. This weighting achieves superior results compared with combining shape and color in an early fusion or late fusion approach. It is worth noting that the authors incorporate both top-down and bottom-up saliency in their approach. However, their approach largely ignores the non-salient regions, and thus it may not work as well on different datasets.

Siagian and Itti \cite{itti-scene} showed that rather than using saliency to prune other features, the saliency itself can be used as a feature. The authors use saliency to form both a local feature and a global (gist) feature. Their approach is specific to the Itti model, and may not generalize to other saliency models. They report encouraging results for the application of scene recognition. 

Kienzle et al. \cite{kienzle} learn a spatio-temporal interest point detector from eye tracking data. They then use this trained interest point detector to classify videos of human activities, and achieve a greater classification accuracy than when using other detectors.

Most closely related to our work, Vig et al. \cite{vigECCV} find that weighting histogram based features by spatio-temporal saliency improves the performance of action recognition in videos. They also use context as an additional feature, however their results show that their saliency model does no better than a Gaussian center bias model. They do show that using eye tracking data to weight features achieves higher classification accuracy than using Gaussian model, meaning that there is much room for improvement.

\subsection{Procedure for Evaluation of Saliency Weighting and Pruning}
\begin{figure}[tb]
	\centering
	\subfloat[Badminton]{	\includegraphics[width=0.23\textwidth]{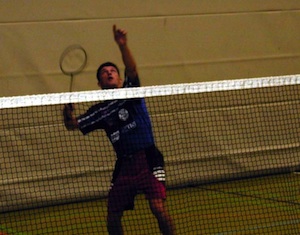}}                
	\subfloat[Bocce ball]{	\includegraphics[width=0.23\textwidth]{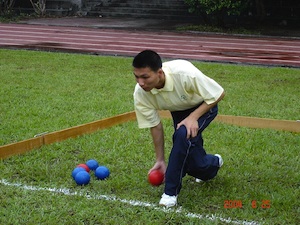}}
	\subfloat[Croquet]{	\includegraphics[width=0.23\textwidth]{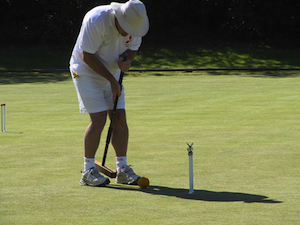}}
	\subfloat[Polo]{	\includegraphics[width=0.23\textwidth]{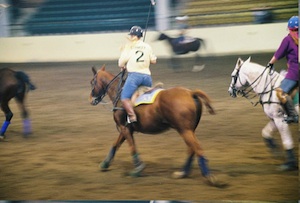}} \\ 
	\subfloat[Rock Climbing]{	\includegraphics[width=0.23\textwidth]{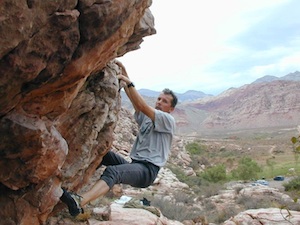}}                
	\subfloat[Rowing]{	\includegraphics[width=0.23\textwidth]{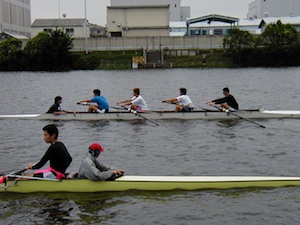}}
	\subfloat[Sailing]{	\includegraphics[width=0.23\textwidth]{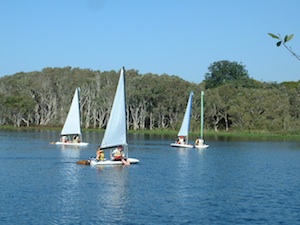}}
	\subfloat[Snowboarding]{	\includegraphics[width=0.23\textwidth]{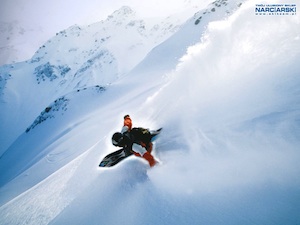}}
	\caption{The UIUC sports dataset \cite{jiaLi-db} consists of 8 categories. Most of the images have clear salient objects or regions.}
	\label{fig:db}
\end{figure}

\begin{figure}[tb]
\begin{tabular}{ m{2.1cm} m{2.4cm} m{2.4cm} m{2.4cm} m{2.4cm}}
  (a) Image 			& \includegraphics[width=\sw]{badminton} & \includegraphics[width=\sw]{bocce.jpg}
  				& \includegraphics[width=\sw]{croquet} & \includegraphics[width=\sw]{polo.jpg}  \\
  (b) Itti \cite{itti98} 		& \includegraphics[width=\sw]{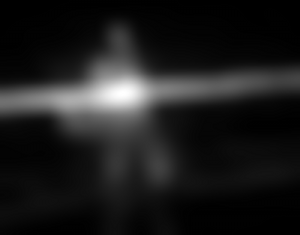} & \includegraphics[width=\sw]{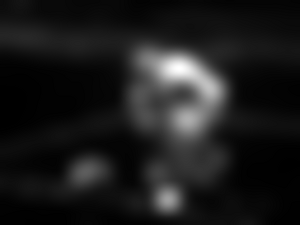}
  				& \includegraphics[width=\sw]{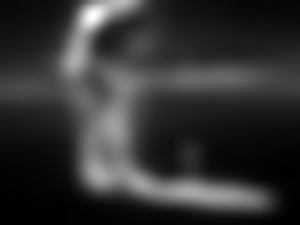} & \includegraphics[width=\sw]{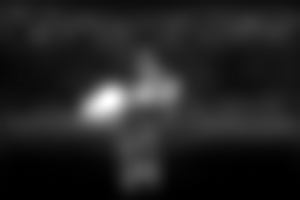}  \\
  (c) GBVS \cite{gbvs} 	& \includegraphics[width=\sw]{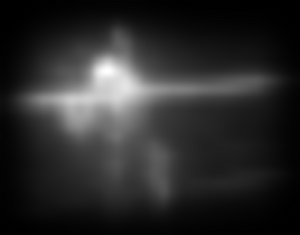} & \includegraphics[width=\sw]{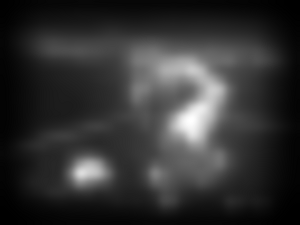}
  				& \includegraphics[width=\sw]{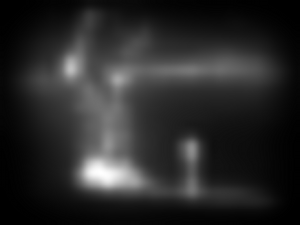} & \includegraphics[width=\sw]{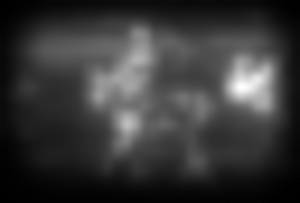}  \\
  (d) RARE \cite{rare} 	& \includegraphics[width=\sw]{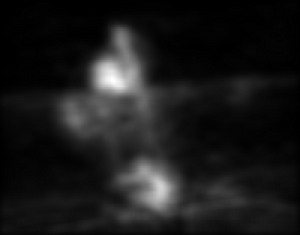} & \includegraphics[width=\sw]{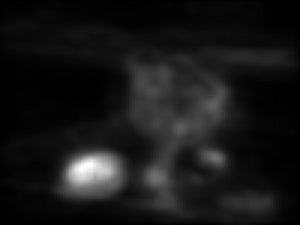}
  				& \includegraphics[width=\sw]{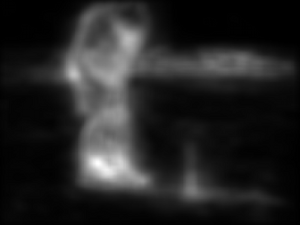} & \includegraphics[width=\sw]{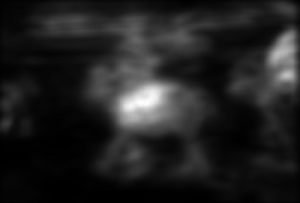}  \\
  (e) AWS \cite{aws}	& \includegraphics[width=\sw]{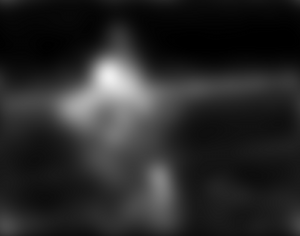} & \includegraphics[width=\sw]{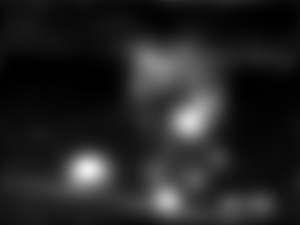}
  				& \includegraphics[width=\sw]{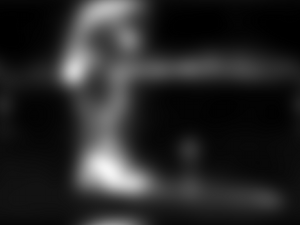} & \includegraphics[width=\sw]{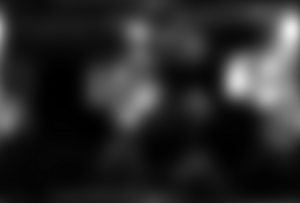}  \\
\end{tabular}
\caption{Comparison of saliency maps on UIUC sports dataset. Some of the models do not perform well on certain images because they will label shadows or portions of the background as salient.}
	\label{fig:smaps}
\end{figure}

To the best of our knowledge there does not exist an evaluation of techniques for using saliency to prune descriptors. Is it best to weight descriptors? Or prune them? Furthermore, what is the best model to use? Does the performance vary significantly from model to model?

To answer these questions, we consider the basic bag of features framework with spatial pyramid matching (SPM) \cite{lazebnik}. In the bag of features framework, an image is described in terms of the relative occurrence of quantized local image features. First, SIFT features are computed in a dense grid at multiple scales. From all SIFT patches in the training set, vector quantization is used to form a codebook of $m$ representative codewords. A feature vector is formed by assigning patches to the nearest codeword and forming a histogram of the occurrence of codewords. This histogram is calculated at different divisions of the image in a pyramid structure. All of the histograms are concatenated together to form a feature vector for a single image. Finally, we use a support vector machine with a $\chi^2$ kernel for classification. 

In this paper we first evaluate two methods for incorporating saliency into this bag of words framework:
\begin{enumerate}
  \item \textbf{Saliency Thresholding} In this approach we take only the $n$ features with corresponding highest saliency values. $n$ can be seen as a parameter that affects the tradeoff between computation time and classification accuracy.

  \item \textbf{Saliency Weighting} In this approach, we first normalize the saliency map so that the maximum is 1. Then we weight each feature by its corresponding saliency value. Unlike the first approach, this method does not yield any benefit in terms of computation time. However, we test it to see if classification accuracy improves.
\end{enumerate}

These two approaches are representative of the general strategy for incorporating saliency in scene recogntion.

\subsection{Results}

\begin{figure}[tb]
	\centering
	\subfloat[Pruning with top $n$ points]{\input{pruneresults}} 
	\subfloat[Weight by saliency]{\includegraphics[width=0.5\textwidth]{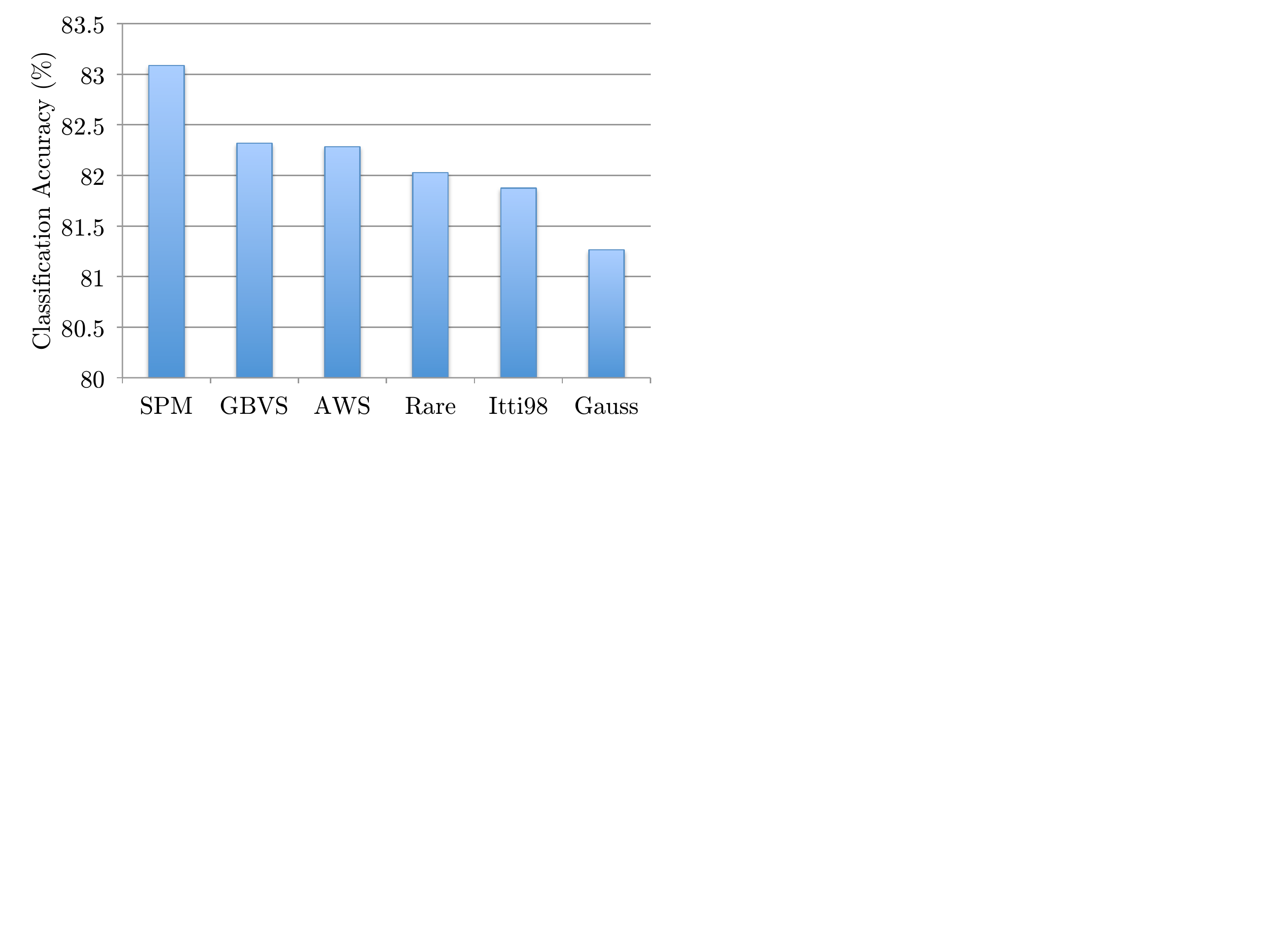}}
	\caption{Saliency weighting and pruning on the UIUC sports dataset with 30 training samples. Pruning by saliency (a) can achieve good accuracy while discarding many of the features. Weighting by saliency (b) offers no advantages in this dataset over the baseline SPM approach \cite{lazebnik}.}
	\label{fig:prune}
\end{figure}

We evaluate saliency pruning and weighting on the UIUC sports dataset \cite{jiaLi-db} (Figure~\ref{fig:db}). The dataset consists of 8 categories: rowing, polo, snowboarding, sailing, badminton, bocce ball, croquet, and rock climbing. Each category contains between 137 and 250 images. All of the images include a clear salient object, namely the person(s) playing the sport.

Here we discuss the parameters used for the bag of words approach. First the images are resized to a common height of 480 pixels. The width is allowed to vary so that the aspect ratio of the image does not change. SIFT features are calculated on a dense grid of two pixels at scales 4, 6, 8, and 10. In forming our codebook we use 600 codewords. We run k-means five times with different initial clusters, and take the codebook with the minimal energy. Spatial pyramid matching with 3 levels is used to create a feature vector for an image. We use the VLFeat library \cite{vlfeat} to construct the features, and LibLinear \cite{liblinear} for classification.

We consider five saliency models that are representative of bottom-up models (Figure~\ref{fig:smaps}). The Itti98 model is a classical approach that implements the center surround feature response and the feature integration theory \cite{itti98}. We use the implementation provided by the GBVS package \cite{gbvs}. Graph based visual saliency (GBVS) \cite{gbvs} is an extension of the Itti98 model that uses random walks on a graph to determine saliency. The RARE \cite{rare} model predicts saliency using both a local and global rarity function with low and mid-level features. Finally, Adaptive Whitening Saliency (AWS) \cite{aws} is a more recent model that has been found to achieve the highest performance in a recent survey paper \cite{borji-tip}. We also consider a simple Gaussian blob as a model. The photographer's bias tends to place objects of interest near the center of the image, so the Gaussian blob models this bias.

To test the classification performance of these models, we train on 30 random images and test on the remaining images. We repeat this five times and average the results. Figure~\ref{fig:prune} shows the results for both pruning descriptors and weighting by saliency. For pruning descriptors the simple gaussian model gives the best performance. This can be attributed to a strong photographer's bias in the dataset. GBVS does well for the same reason: it has a center bias built in. Surprisingly, the AWS model performs poorly in saliency pruning and saliency weighting despite good performance on eye tracking datasets.

\section{Combining Salient and Non-salient Regions for Improved Performance}

\begin{figure}[tb]
	\centering
	\includegraphics[width=1\textwidth]{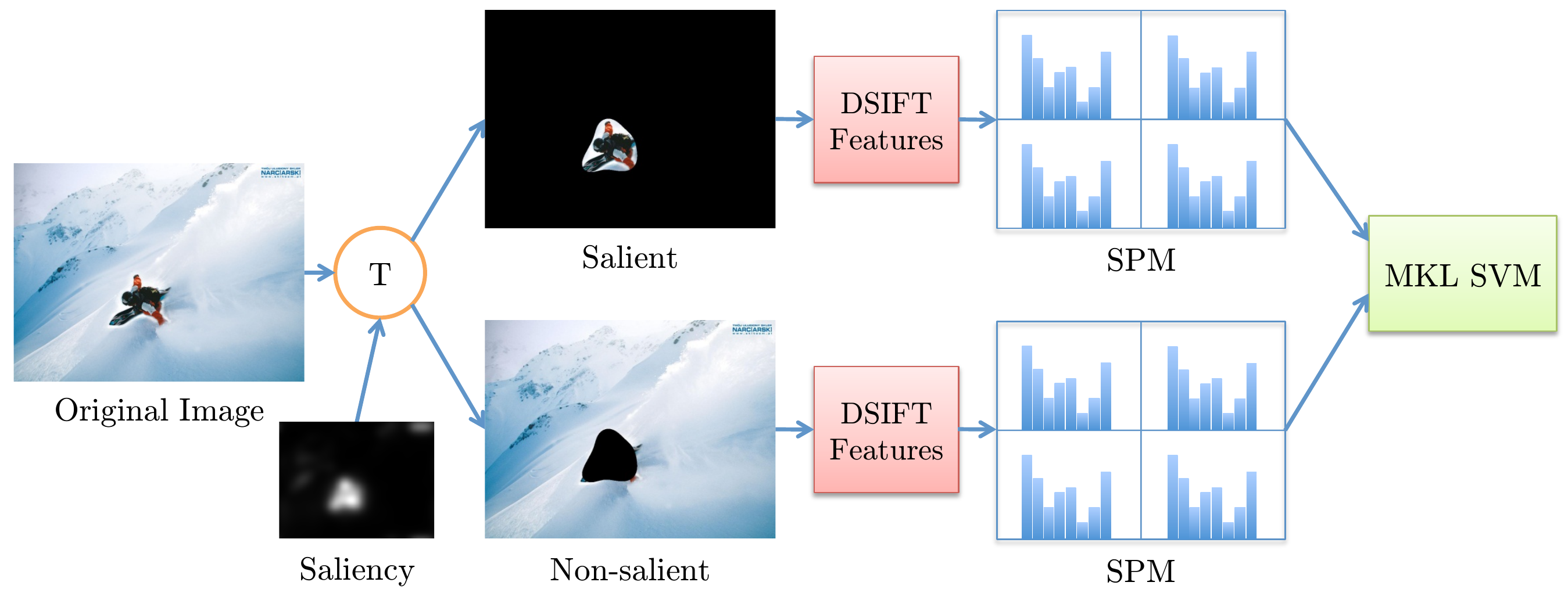}
	\caption{Overview of the proposed saliency weighted approach. $T$ represents the combined operation of thresholding the saliency map and multiplying with the original image to obtain salient and non-salient regions.} 
	\label{fig:approach}
\end{figure}

Descriptor pruning remains popular because it can yield significant computational savings without a significant loss of classification performance. However, the pruning based methods may discard a significant amount of useful information that could be useful to encode the context of the scene. For example, in the UIUC sports dataset the salient region might be a snowboarder, but the rest of the image (the mountain and the snow) may tell just as much about the action in the image. Thus we propose to combine information from the salient and non-salient areas of the scene to achieve higher classification accuracy.

It is well known in the computer vision community that context can improve object or scene recognition. Galleguillos and Belongie \cite{context-survey} provide an overview of the use of context in computer vision. Here we discuss more recent work. Marat and Itti \cite{itti-context} use HMAX and gist features to encode the object and its context. They find that context significantly improves performance on a synthetic database. Marszaek and Schmid \cite{marsalek} attempt to learn foreground and background, and then suppress background features. In our approach we use a low-level model of the attention mechanism of the human visual system to separate the region of interest from the contextual information. This is both biologically motivated, and may be more computationally efficient than the pure computer vision approaches.


We again use the basic bag of features framework described in Section 2 as our baseline. In our approach, we first compute the saliency of the image. The image is then divided into salient regions and non-salient regions using a simple threshold. Two separate bag of features histograms are formed using spatial pyramids \cite{lazebnik}: one for the salient regions and one for the non-salient regions.

\begin{figure}[t!]
	\centering
	\input{resultscomp}
	\caption{Recognition performance on UIUC sports dataset with different models using our proposed method. The AWS model yields the best performance for all number of training sizes.}
	\label{fig:acccomp}
\end{figure}
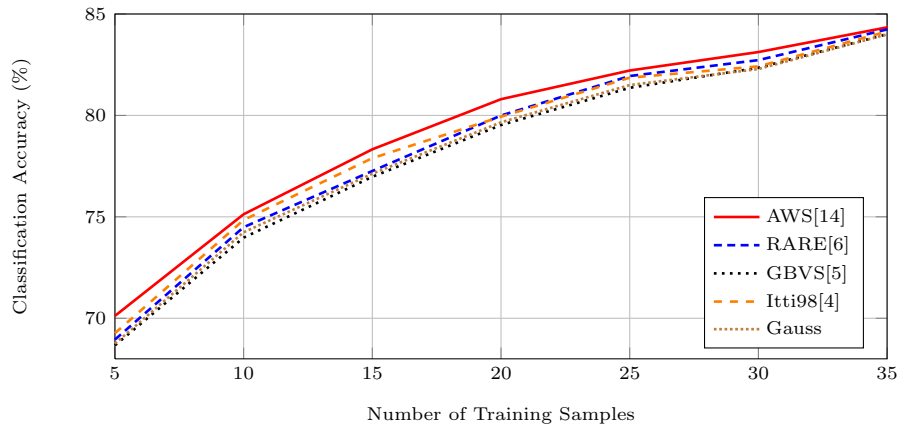

Next we must learn the relative influence of the information in the salient and non-salient regions. We treat each part separately and use multiple kernel learning (MKL) \cite{mkl} to combine the information. Multiple kernel learning attempts to find the best linear combination of SVM kernels. In our case we have 2 kernels: one corresponding to SPM features for the salient areas, and the other corresponding to SPM features for the non-salient areas. Specifically the MKL algorithm finds the optimal $K^*$:

$$K^* = \alpha K_s + (1-\alpha) K_{ns}$$

where $K_s$ is the $\chi^2$ kernel for the salient regions and $K_{ns}$ is the $\chi^2$ kernel for the non-salient regions. The MKL algorithm finds the optimal $\alpha$. In our experiments, we use a version of the Liblinear MKL Matlab library\footnotemark~that has been modified for use with custom kernels.

\footnotetext{By Ming-Hen Tsai. Available at \url{http://www2.csie.ntu.edu.tw/~b95028/software/liblinear-mkl/index.php}}

\subsection{Results}

First we test the classification accuracy of five saliency models in our proposed approach. We train on a number of random training samples and test on the remaining images of the dataset. We run each test five times, using a different random training set each time, and then take the average classification accuracy. For all models we set the saliency threshold to $T=0.5$. We find that the AWS model \cite{aws} achieves the best performance in our method (Figure~\ref{fig:acccomp}). This is consistent with studies showing that AWS performs the best on eye tracking datasets \cite{borji-tip}.

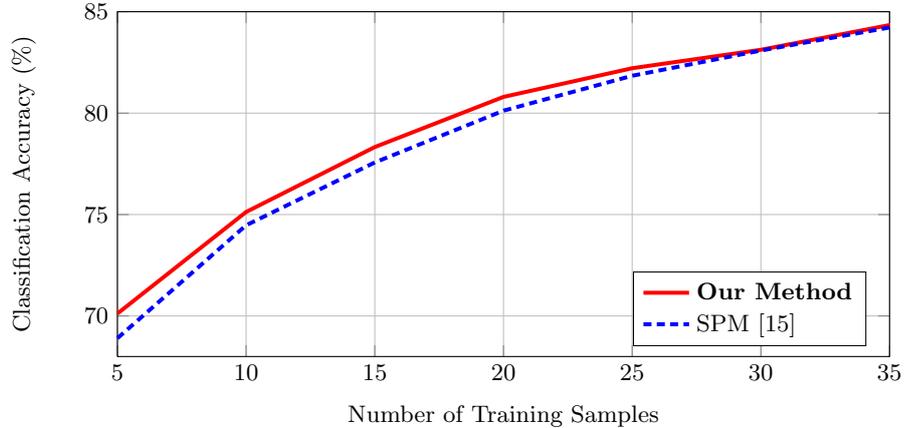
\begin{figure}[t!]
	\centering
	\input{results}
	\caption{Recognition performance on UIUC sports dataset with our method using the AWS model.}
	\label{fig:acc}
\end{figure}

Our proposed method using the AWS model yields higher accuracy than the baseline SPM approach for a small training size (Figure~\ref{fig:acc}). Both approaches use features calculated on the whole image, so it is clear that separating the salient regions from the non-salient regions gives greater discriminative ability.

With a large training size, any benefit from treating the salient regions separately from the non-salient becomes negligible. This may be because the richness of a larger training set cancels out any performance gains. For small training size, the extra information from the non-saliency proves useful. It is also worth noting that many times, the saliency models do not produce reasonable maps. What should not be salient is predicted as salient, and thus the distinction between salient and non-salient in these cases is weak. If the saliency model is improved, the performance of our method will also improve.


\section{Discussion and Conclusion}

Our improved approach can be seen as modeling the concept of gaze and periphery. While it is true that the human visual system will fixate on certain salient regions in an image, the human visual system also has the ability to recognize in the periphery \cite{perif}. The salient region features could represent the gaze regions, whereas the non-salient region features could represent the peripheral gist.

It must be noted that the current saliency models fail on many images in the dataset. Although we have not performed any eye tracking experiments, most of the images in the dataset have clear salient objects (usually the human performing the action). It may be that the newer saliency models are beginning to over-fit the standard eye tracking datasets. The most widely used database to verify model performance consists of only 120 images \cite{aim}. These 120 images may not be representative of the real world. Additionally, Tatler et al. noted that the eye tracking experiments themselves are flawed and are not representative of how we see the world \cite{tatler2011}. Finally, in this application it may be appropriate to incorporate top-down knowledge into the saliency maps. Nevertheless, we have shown that even a suboptimal saliency model with our approach yields improvement over the baseline bag-of-words approach for a small number of training samples.

In future work we would like to apply this approach on a wider variety of problem domains, such as object recognition (Caltech 101, and Pascal VOC datasets), or other scene recognition datasets (MIT Indoor scenes, and Scene15). The proposed method works well on the UIUC sports dataset because in many cases there is a clear salient object. In other problem domains, there may not be a clear salient object or objects to pick out. Additionally, it would be useful to obtain ground truth eye-tracking on the UIUC dataset both to see which model is the most accurate and to obtain an upper bound on the performance of our classification approach.

In this paper, we have analyzed several different approaches that incorporate saliency in the application of scene recognition. We show that although pruning descriptors by saliency reduces computation time, it offers no performance benefit in terms of classification accuracy. For this dataset, weighting by saliency also offers no performance advantage over the baseline bag of words model. However, we propose a method that uses salient and non-salient regions of an image separately for classification. Compared with the baseline SPM approach, our results show improved performance with an appropriate saliency map for a small training set on the UIUC sports dataset.

\bibliographystyle{ieeetr}
\vspace{-10pt}
\bibliography{bib}

\end{document}

%% file: pruneresults.tex
%
%
%
%

\definecolor{ppt-blue}{rgb}{0.356,0.588,0.772}
\definecolor{ppt-red}{rgb}{0.753,0.313,0.301}
\definecolor{ppt-green}{rgb}{0.607,0.733,0.349}
\definecolor{ppt-purple}{rgb}{0.501,0.392,0.635}
\definecolor{ppt-turq}{rgb}{0.294,0.674,0.776}
\definecolor{ppt-orange}{rgb}{0.968,0.588,0.274}

\begin{tikzpicture}

\begin{axis}[%
grid = major,
width=1.5in,
height=1.5in,
scale only axis,
xmin=10, xmax=100,
ymin=45, ymax=85,
xlabel=Percent of Descriptors Kept (\%),
ylabel=Classification Accuracy (\%),
y label style={at={(0.15,0.5)}},
legend style={nodes=right},
legend pos= south east,
style={font=\scriptsize}]

\addplot [
color=black,
dotted, line width = 1
]
table{
10 61.035317
20 67.093550
30 71.246781
40 74.099509
50 75.244986
60 76.790735
70 77.933823
80 79.652354
90 80.987816
100 82.330317
};
\addlegendentry{Gauss};

\addplot [
color=red,
dashed, line width = 1
]
table{
10 61.993697
20 68.484990
30 70.595546
40 72.781605
50 74.354295
60 75.238024
70 76.480077
80 78.437157
90 81.122025
100 82.330317
};
\addlegendentry{GBVS \cite{gbvs}};

\addplot [
color=orange,
densely dotted, line width = 1
]
table{
10 50.391835
20 59.783864
30 65.683379
40 69.291633
50 71.826386
60 74.643920
70 76.662894
80 79.429915
90 81.515546
100 83.158641
};
\addlegendentry{RARE \cite{rare}};

\addplot [
color=blue,
densely dashed, line width = 1
]
table{
10 52.880313
20 58.687853
30 63.887456
40 67.923147
50 72.005252
60 73.895021
70 76.276877
80 78.265258
90 80.624455
100 82.330317
};
\addlegendentry{Itti98 \cite{itti98}};

\addplot [
color=brown,
solid, line width = 1
]
table{
10 49.324970
20 57.548038
30 62.743053
40 65.425249
50 69.871247
60 73.407459
70 76.100523
80 78.668228
90 81.169025
100 82.330317
};
\addlegendentry{AWS \cite{aws}};

\end{axis}
\end{tikzpicture}%

%% file: resultscomp.tex
%
%
%
%
\begin{tikzpicture}

\begin{axis}[%
grid = major,
width=4in,
height=1.8in,
scale only axis,
xmin=5, xmax=35,
ymin=68, ymax=85,
xlabel=Number of Training Samples,
ylabel=Classification Accuracy (\%),
legend style={nodes=right},
legend pos= south east,
style={font=\scriptsize}]

\addplot [
color=red,
solid, line width = 1
]
table{
5 70.116662
10 75.126540
15 78.328346
20 80.799773
25 82.213761
30 83.123518
35 84.341642
40 84.851474
45 85.657079
50 86.436655
55 87.065829
60 87.741990
};
\addlegendentry{AWS\cite{aws}};

\addplot [
color=blue,
densely dashed, line width = 1
]
table{
5 68.948493
10 74.490051
15 77.259335
20 79.995212
25 81.942377
30 82.726538
35 84.239799
40 84.753905
45 85.410815
50 86.504212
55 87.066231
60 87.793900
};
\addlegendentry{RARE\cite{rare}};

\addplot [
color=black,
dotted, line width = 1
]
table{
5 68.662460
10 73.973225
15 76.972298
20 79.525415
25 81.360662
30 82.348466
35 84.002351
40 84.198944
45 84.968494
50 85.729910
55 86.210642
60 86.872394
};
\addlegendentry{GBVS\cite{gbvs}};

\addplot [
color=orange,
dashed, line width = 1
]
table{
5 69.261142
10 74.843479
15 77.892650
20 79.936707
25 81.859184
30 82.418627
35 84.135197
40 84.887100
45 85.433683
50 86.113185
55 86.791748
60 87.334020
};
\addlegendentry{Itti98\cite{itti98}};

\addplot [
color=brown,
densely dotted, line width = 1
]
table{
5 68.746719
10 74.239711
15 77.127821
20 79.660465
25 81.496084
30 82.286958
35 83.996561
40 84.213645
45 84.479292
50 85.631062
55 86.619987
60 86.818464
};
\addlegendentry{Gauss};


\end{axis}
\end{tikzpicture}%

%% file: results.tex
%
%
%
%
\begin{tikzpicture}

\begin{axis}[%
grid = major,
width=4in,
height=1.8in,
scale only axis,
xmin=5, xmax=35,
ymin=68, ymax=85,
xlabel=Number of Training Samples,
ylabel=Classification Accuracy (\%),
legend style={nodes=right},
legend pos= south east]

\addplot [
color=red,
solid, line width = 1.5
]
table{
5 70.116662
10 75.126540
15 78.328346
20 80.799773
25 82.213761
30 83.123518
35 84.341642
40 84.851474
45 85.657079
50 86.436655
55 87.065829
60 87.741990
};
\addlegendentry{\textbf{Our Method}};

\addplot [
color=blue,
densely dashed, line width = 1.5
]
table{
5 68.901696
10 74.471422
15 77.568194
20 80.120676
25 81.841728
30 83.085285
35 84.221427
40 84.812081
45 85.507361
50 86.304246
55 87.040672
60 87.523262
};
\addlegendentry{SPM \cite{lazebnik}};



\end{axis}
\end{tikzpicture}%